\documentclass[journal]{IEEEtran}
\usepackage[noadjust]{cite}
\usepackage{multicol}
\usepackage{graphicx}
\usepackage{subcaption}
\usepackage{multirow}
\usepackage{mathtools}
\usepackage{breakcites}

\ifCLASSINFOpdf
\else
\fi
\begin{document}
%
\title{NU-LiteNet: Mobile Landmark Recognition using Convolutional Neural
Networks}
%
%
%

\author{Chakkrit~Termritthikun, Surachet~Kanprachar, Paisarn~Muneesawang
\IEEEcompsocitemizethanks{\IEEEcompsocthanksitem 

C. Termritthikun is with the Department of Electrical and Computer Engineering, Faculty of Engineering, Naresuan University, Phitsanulok, Thailand. 
\protect\\ E-mail: chakkritt60@email.nu.ac.th.
\IEEEcompsocthanksitem S. Kanprachar is with the Department of Electrical and Computer Engineering, Faculty of Engineering, Naresuan University, Phitsanulok, Thailand.
\protect\\ E-mail: surachetka@nu.ac.th.
\IEEEcompsocthanksitem P. Muneesawang is with the Department of Electrical and Computer Engineering, Faculty of Engineering, Naresuan University, Phitsanulok, Thailand.
\protect\\ E-mail: paisarnmu@nu.ac.th.}
\thanks{Manuscript received April 19, 2005; revised August 26, 2015.}}

\maketitle

\begin{abstract}
The growth of high-performance mobile devices has resulted in more
research into on-device image recognition. The research problems are
the latency and accuracy of automatic recognition, which remain obstacles
to its real-world usage. Although the recently developed deep neural
networks can achieve accuracy comparable to that of a human user,
some of them still lack the necessary latency. This paper describes
the development of the architecture of a new convolutional neural
network model, NU-LiteNet. For this, SqueezeNet was developed to reduce
the model size to a degree suitable for smartphones. The model size
of NU-LiteNet is therefore 2.6 times smaller than that of SqueezeNet.
The recognition accuracy of NU-LiteNet also compared favorably with
other recently developed deep neural networks, when experiments were
conducted on two standard landmark databases.
\end{abstract}


%
\IEEEpeerreviewmaketitle

\section{Introduction}
%
%
%
%
\IEEEPARstart{L}{andmark} recognition is an important feature for
tourists who visit important places. A tourist can use a smartphone
that installs a landmark-recognition application for retrieval of
information about a place, such as the names of landmarks, the history,
events that are currently taking place, and opening times of shows.
This process involves taking a picture of a landmark and letting the
application software retrieve the relevant information. This effective
mobile interface has created new trends for the tourist industry,
mobile shopping, and other e-commerce applications.

In the past, landmark recognition \cite{1,2,3,4,6} utilized the
capability of computers. These computing devices can cope with the
large size of databases and the computational complexity, with sufficient
resources to operate the application. However, the major problems
are the accuracy of recognition and the long processing time when
the applications are running on other mobile devices. These may be
because of the utilization of recognition methods such as the scale
invariant feature transform (SIFT), scalable vocabulary tree (SVT),
and geometric verification (GV). Some of these methods have been studied
extensively in the past because of their exceptional performance.
However, their high accuracy results in long processing times.

The application of machine learning models for landmark recognition
has encountered various problems in practice. Landmark recognition
needs a large amount of training with a dataset to obtain an effective
machine learning model. This model is then utilized by the recognition
program. The size of the model obtained is usually great, and thus
requires a long time for processing. The image processing and recognition
are therefore usually done on the server computer. The picture is
taken by the smartphone users and sent to the server for recognition,
after which the result is sent to the smartphone. Moreover, the smartphone
has to be connected to the internet to perform the recognition function;
it cannot be performed in off-line mode. To solve this problem, the
application needs to embed the machine leaning model into the smartphone
and perform on-device recognition. However, this large model cannot
fit into the smartphone because of the latter\textquoteright s limited
memory space, and so its size has to be reduced. One method for doing
so is the application of a convolutional neural network (CNN). This
has been recently studied with a view to extending the CPU and GPU
modules to achieve high-performance image recognition.

CNN has received much attention for image recognition, object detection,
and image description. For the ImageNet Large Scale Visual Recognition
Challenge (ILSVRC), new models have been developed, and are more effective
than the previous models. Such models are AlexNet \cite{8}, GoogLeNet
\cite{9}, VGG \cite{10}, and ResNet \cite{11}, which were the winners
in 2012\textendash 2015. These competitions have stimulated progress
in the development of research on image recognition, and the CNN models
are the most effective examples of machine learning at present.

As described in \cite{12}, AlexNet \cite{8}, the winner of ILSVRC 2012,
was applied to a large-scale social image collection (500 classes
of 2 million images), and compared with the Bag-of-Word (BoW) method
using a SIFT descriptor. It was shown that CNN could attain 23.88\%
recognition accuracy, while the BoW method only reached 9.5\%. This
result indicated that CNN is more effective for image recognition
than BoW methods. As described in \cite{13}, AlexNet was modified by
reducing the parameters of SqueezeNet by up to 50 times, which resulted
in a \textquotedblleft lite\textquotedblright{} version of CNN. The
structure of SqueezeNet contains two parts: (1) A Squeeze block, which
implements the convolution layer with a $1\times1$ filter, and
(2) an Expand block, which implements the convolution layer with $1\times1$
and $3\times3$ filters. The Squeeze block reduces the data dimension,
while the Expand block is effective in analyzing data. The reduced-size
version of CNN can still maintain the same level of recognition accuracy
as AlexNet.

GoogLeNet was developed by Google and was the winner of ILSVRC 2014. A defining feature of GoogLeNet is its inception module, with the ability to analyze data accurately. The network consists of a convolution layer with $1\times1$, $3\times3$, and $5\times5$ filters. It uses a convo-lution layer with a $1\times1$ filter to reduce the data dimensions.
GoogLeNet can reduce the model size up to 4.8 times more than AlexNet. The architecture of GoogLeNet
includes nine inception modules arranged in a cascad-ing manner, which increases performance in terms of
recognition accuracy. However, this structure also
increases the time required to train the network to about three times that of AlexNet.

For this paper, we adopt the idea for the development of SqueezeNet, which consists of a Squeeze block and Expand block. The improvement consists of the inclusion of a convolution layer with $5\times5$ and $7\times7$ filters to enable the Expand block to cope with the analysis of complex image content. It is also proposed to conduct the Squeeze block in order to reduce the data dimensions. The newly proposed network, NU-LiteNet, can achieve high
recognition accuracy as well as reduced processing time, by using CNN models of the minimum possible size. This makes on-device processing possible, particularly for the landmark recognition facility on smartphones.

\section{Convolutional Neural Networks}
The convolutional neural network (CNN) has a structure the same as that of a normal neural network. It is
classified as a feed-forward neural network, which
consists of a convolution layer, pooling layer, and fully connected layer. At least three of these layers are stacked on a network for learning and classifying data. These
layers, as well as the input layer, are placed in the following order:

The input layer is a layer that contains an image
dataset for training and testing. The image data is in RGB color space and the image size depends on the selected network model. For example, the network model that utilizes an image width of 256 pixels and height of 256 pixels will have data for one image at $[256\times256\times3]$, where 3 is the number of color channels.

The convolution layer is the layer that operates for the multiplication of each pixel with filter coefficients. The operation starts at location (0,0) of the data, and moves by one pixel (stride 1) each time from left to right and top to bottom until all pixels are covered. This process will
result in the creation of an activation map. For example, given that the size of the image data is $[224\times224\times3]$ and there are a total of 96 filters, each of which has a size of $3\times3$, the resulting activation map will be $[111\times111\times3]$ when the filter moves by two pixels (stride 2) each time.

The pooling layer comes after the convolution layer. Its main function is to reduce the dimensions of the data
representation, which will reduce the number of parameters and calculations in the next layer. The max pooling is the function that perform this task. For example, in order to reduce data of size $[111\times111\times3]$ to half that size
(i.e., $[55\times55\times3]$), a filter of size $3\times3$ and stride 2 are needed.

The last layer is the fully connected layer. Its main function is to convert the output data to one dimension.
The CNN can be developed to learn a dataset by
increasing the number of hidden layers in order to
increase leaning capability. The network will divide
image data into sub-images, each of which is analysed for features such as color shape and texture. These features will be used for the prediction patterns for image
classification.

\begin{figure}[b]
  \centering
  \includegraphics[scale=0.3]{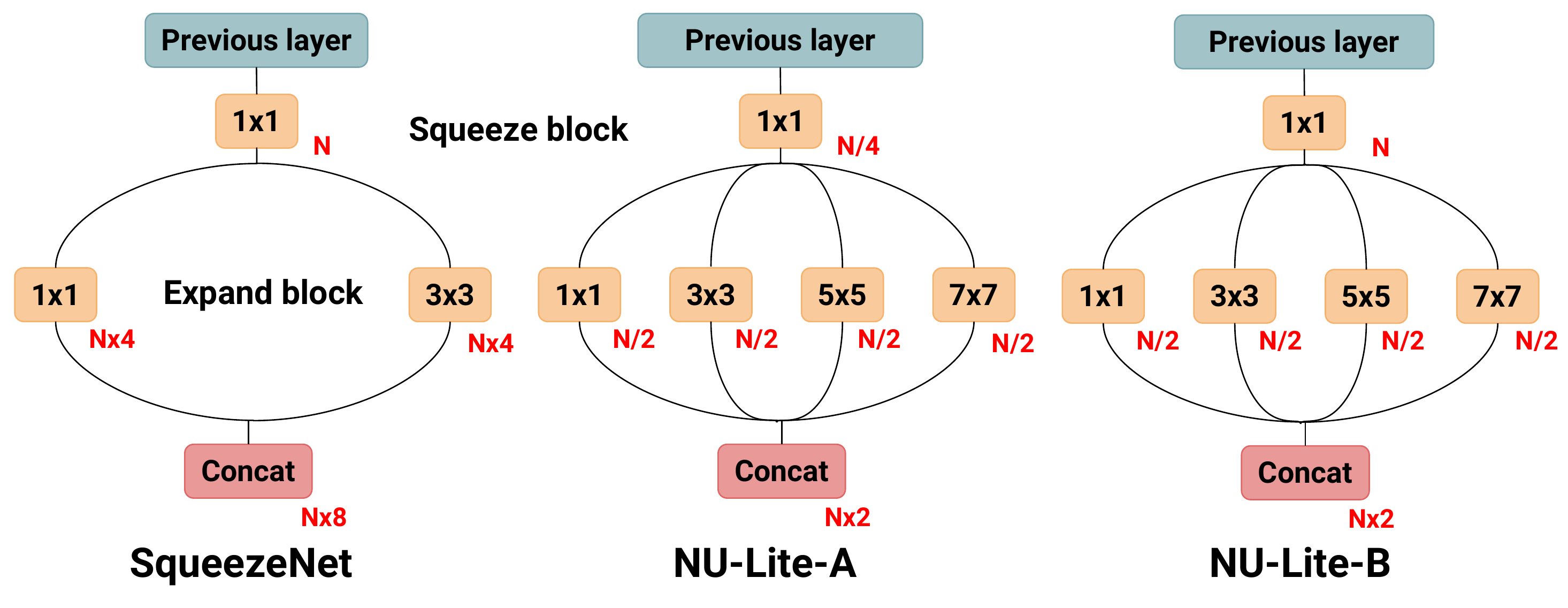}
  \caption{Squeeze block and Expand block of NU-LiteNet-A and NU-LiteNet-B compared with SqueezeNet}\label{fig1}
\end{figure}

\section{NU-LiteNet}

This section presents the development of two types of network architecture for CNNs: NU-LiteNet-A and NU-LiteNet-B.

\subsection{Added $5\times5$ and $7\times7$ Convolution}

Considering SqueezeNet's Expand blocks \cite{13}, SqueezeNet choose the use of small filter, such as $1\times1$ and $3\times3$ convolution, to detect smaller objects. Another reason for using a small filter comes from the design of the model, for the size of the parameter is small and the processing time is minimal. As a result of this, SqueezeNet's accuracy is not as high compared to GoogLeNet \cite{9}, but has the same accuracy level as AlexNet. \cite{8} In this paper, we choose the use of large convolution filter, such as $5\times5$ and $7\times7$ added to the Expand blocks in order to enhance the accuracy, just as the Inception module of the GoogLeNet \cite{9}. The use of a large filter to detect objects similarly to the small filter, but the difference is that the Large filter helps to identify or confirm the central position of the object. When the data from the small filter and large filter are concatenated, the model can confirm the position of the desired object as shown in \cite{14,15}. For this reason, the model efficiency has greater accuracy. However, increase in the Large filter $5\times5$ and $7\times7$ convolution expand blocks, results into increase in the processing time and the number of parameters. Therefore, there’s need to reduce the size and depth of model of SqueezeNet because of its large filter expand blocks. So that the processing time and the number of parameters with the appropriate size and applications can be properly processed on smartphones.

\subsection{NU-LiteNet-A}

NU-LiteNet-A was developed by changing SqueezeNet, which has the Squeeze and Expand blocks, as shown in Fig. 1(a). It introduces $5\times5$ convolution and $7\times7$
convolution into the Expand block, as shown in Fig. 1(b). If N is the number of channels (depth) of the
previous layer, NU-LiteNet-A will reduce N in the $1\times1$ convolution or Squeeze block by one fourth (i.e., $\frac{N}{4}$) of the previous layer. Next, it will increase N in the
Expand block to double (i.e., $\frac{N}{2}$) that of the Squeeze block. As a result, the number of channels will be
increased to double (i.e., $N\times2$) that of the previous layer after the Expand block. The details of NU-LiteNet-A are summarized in Table 1.

\subsection{NU-LiteNet-B}

NU-LiteNet-B changes the structure of NU-LiteNet-A by changing the amount of depth, N, of the Squeeze block to the same of that of the previous layer. This
corresponds to the structure of SqueezeNet as shown in Fig. 1(c). In this structure, the Expand block will receive an amount of depth, N, equal to that of the previous layer. This increases the effectiveness of the net-work for data analysis, but will also increase the number of parameters and thus require a longer processing time. The details of NU-LiteNet-B are summarized in Table 1.

\begin{figure}[tbh]
  \centering
  \includegraphics[scale=0.15]{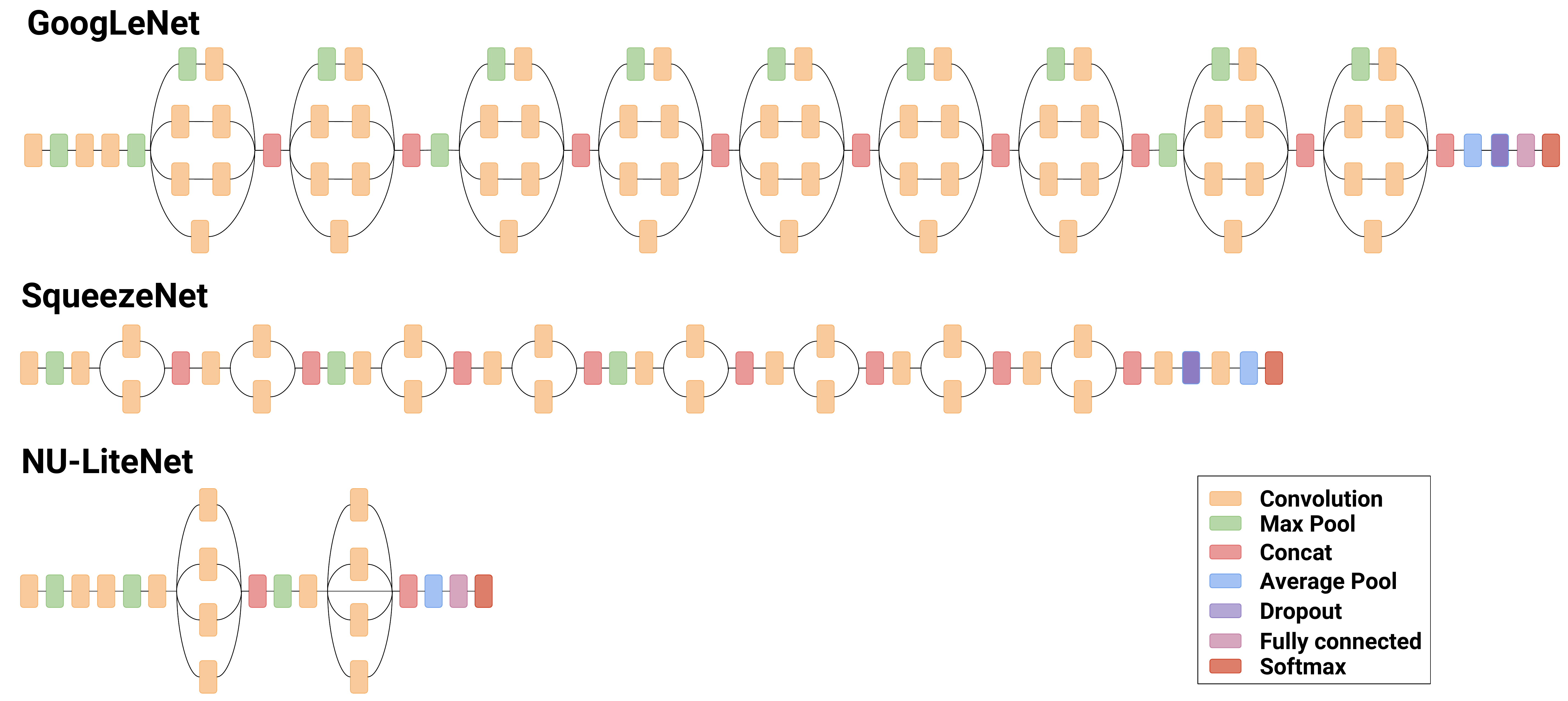}
  \caption{Architecture of NU-LiteNet.}\label{fig2}
\end{figure}

\begin{table}[tbh]
    \centering{}\caption{NU-LiteNet-A and NU-LiteNet-B}
    \begin{tabular}{|c|c|c|c|}
      \hline
      layer name  & output size  & NU-LiteNet-A  & NU-LiteNet-B\tabularnewline
      \hline
      \hline
      Input  & 224x224  & \multicolumn{2}{c|}{-}\tabularnewline
      \hline
      Convolution 1  & 113x113  & \multicolumn{2}{c|}{5x5, 64, stride 2, pad 3}\tabularnewline
      \hline
      Pooling 1  & 56x56  & \multicolumn{2}{c|}{max pool, 3x3, stride 2}\tabularnewline
      \hline
    Convolution 2  & 56x56  & \multicolumn{2}{c|}{1x1, 64, stride 2}\tabularnewline
    \hline
    Convolution 3  & 56x56  & \multicolumn{2}{c|}{3x3, 64, stride 1, pad 1}\tabularnewline
    \hline
    Pooling 2  & 28x28  & \multicolumn{2}{c|}{max pool, 3x3, stride 2}\tabularnewline
    \hline
    NU-Lite-Block 1  & 28x28  & {[}Block-A{]}, 128  & {[}Block-B{]}, 128\tabularnewline
    \hline
    Pooling 3  & 14x14  & \multicolumn{2}{c|}{max pool, 3x3, stride 2}\tabularnewline
    \hline
    NU-Lite-Block 2  & 14x14  & {[}Block-A{]}, 256  & {[}Block-B{]}, 256\tabularnewline
    \hline
    Pooling 4  & 1x1  & \multicolumn{2}{c|}{average pool}\tabularnewline
    \hline
    Fully connected  & 50  & \multicolumn{2}{c|}{softmax}\tabularnewline
    \hline
    \end{tabular}
\end{table}

\subsection{Completed Network structures}

The complete architectures of NU-LiteNet-A and NU-LiteNet-B are shown in Fig. 2. The proposal is to cut the number of layers and include an Expand block. NU-LiteNet-A and NU-LiteNet-B have only two modules each, and the number of channels (depth) is $N=256$ channels. This is because the experimental data (shown in Section 4) has only 50 classes. If the amount of depth is increased, the network will have a large number of parameters and require a longer processing time. There-fore, the design of the network has to consider the number of parameters and the processing time that can be applied effectively on smartphones

This design is suitable for processing in a smartphone. The aim is to obtain a network of high effectiveness that is the same as other state-of-the-art CNN models, while keeping the processing time to a minimum. In Fig. 2, GoogLeNet is shown in comparison with the proposed network architecture. GoogLeNet has nine modules, whereas the proposed network has only two modules, which will reduce processing time and model size.

\section{Experimental result}
In the experiment, we trained the networks with a high-performance computing (HPC) unit. It had the follow-ing specifications: Intel(R) Xeon(R) E5-2683 v3 @ 2.00GHz 56 Core CPU, 64 GB RAM, and NVIDIA Tesla K80 GPU. The operating system was Ubuntu Server 14.04.5. For testing, we used a smartphone with the fol-lowing specifications: Samsung Exynos Octa 7580 @ 1.6 GHz 8 Core CPU and 3 GB RAM, working on Android 6.0.1.

\subsection{Databases}

The experimental data were obtained from two stand-ard landmark datasets. The first set was of Singapore landmarks \cite{2}, and consisted of 50 landmarks (4,060 images) some of which are shown in Fig.3 (a), the im-portant places in Singapore that are popular with tour-ists. The second dataset was the Paris dataset \cite{16}, which consisted of 12 landmarks (6,412 images) some of which are shown in Fig.3 (b) in Paris, France. For each dataset, images were divided into a training set and testing set, at 90\% and 10\% respectively. The images were resized to $256\times256$ pixels.

\begin{figure}[tbh]
\centering
\begin{subfigure}[b]{0.5\textwidth}
   \centering
   \includegraphics[width=0.85\linewidth]{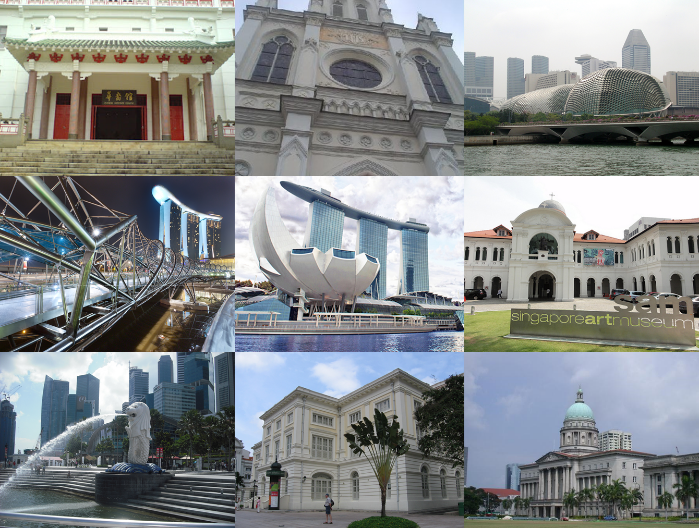}
   \caption{}
   \label{fig:Ng1}
\end{subfigure}

\begin{subfigure}[b]{0.5\textwidth}
      \centering
      \includegraphics[width=0.85\linewidth]{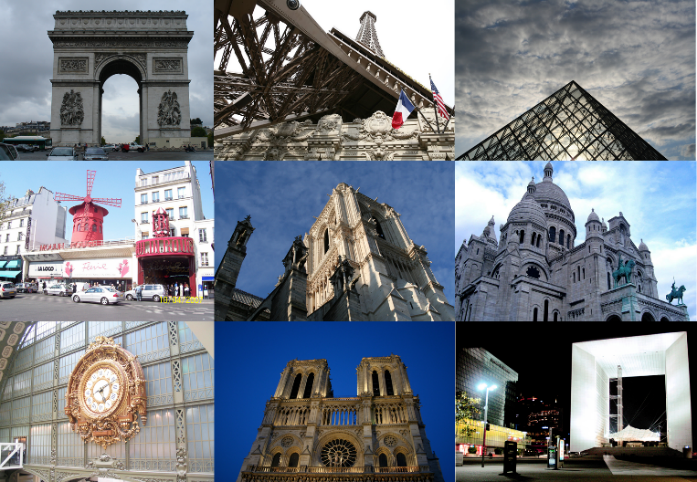}
   \caption{}
   \label{fig:Ng2}
\end{subfigure}

\caption[Sample images ]{(a) Singapore Landmark (b) Paris landmark}
\end{figure}

\subsection{Comparison of NU-LiteNet and other models}

In the experiment, all network models, including AlexNet, GoogLeNet, SqueezeNet, NU-LiteNet-A, and NU-LiteNet-B, were trained from scratch. The Singapore landmarks and Paris dataset were used, and each set was divided into two parts: a training set (90\%) and a testing set (10\%), with 10-fold cross-variation. The hyperparameters for NU-LiteNet-A and NU-LiteNet-B were as follows. Solver: Stochastic Gradient Descent (SGD) \cite{17}; Momentum: 0.9; Mini-batch size: 128; Learn-ing rate: 0.1; Weight decay: 0.0005; Epoch size: 100.

\begin{table}[tbh]
\centering{}\caption{RECOGNITION ACCURACY OBTAINED BY 10-FOLD CROSS-VALIDATION. NU-LITENET IS COMPARED WITH OTHER MODELS, USING THE SINGAPORE LANDMARK DATASET.}
\begin{tabular}{|c|c|c|c|c|}
\hline
Model  & Params (M)  & top-1 acc. (\%)  & top-5 acc. (\%)\tabularnewline
\hline
\hline
AlexNet & 62.37   & 64.82  & 84.94 \tabularnewline
\hline
GoogLeNet & 6.02   & 70.69  & 88.75 \tabularnewline
\hline
SqueezeNet  & 0.75   & 60.08  & 83.24 \tabularnewline
\hline
NU-LiteNet-A  & 0.28   & 78.09  & 92.75 \tabularnewline
\hline
NU-LiteNet-B  & 0.94  & 81.15  & 93.96 \tabularnewline
\hline
\end{tabular}
\end{table}

For the training process, we measured the parameters of the networks. The number of parameters indicated the model size. For the testing process, we measured the accuracy using 10-fold cross-validation. The accuracy was measured in terms of the top-1 accuracy as well as the top-5 accuracy.

Table 2 shows the experimental result obtained by 10-fold cross-validation for the Singapore landmark dataset. It can be observed from the result that both versions of NU-LiteNet were more effective for landmark recognition at top-1 accuracy as well as top-5 accuracy than AlexNet, GoogLeNet, and SqueezeNet. The accuracy was higher than that of GoogLeNet by up to 7.4-10.46\%. For the number of parameters, it was discovered that NU-LiteNet-A had the lowest number of parameters: 0.28M. This was 2.5 times lower than that of SqueezeNet.

The experiment results from the Paris dataset showed similar trends to those of the Singapore dataset in terms of recognition accuracy. Both versions of NU-LiteNet gave higher accuracy than the other models. The accuracy was higher than that of GoogLeNet by up to 6.7-9.61\%, as shown in Table 3.

\begin{table}[tbh]
\centering{}\caption{RECOGNITION ACCURACY OBTAINED BY 10-FOLD CROSS-VALIDATION. NU-LITENET IS COMPARED WITH OTHER MODELS, USING THE PARIS DATASET.}
\begin{tabular}{|c|c|c|c|c|}
\hline
Model  & Params (M)  & top-1 acc. (\%)  & top-5 acc. (\%)\tabularnewline
\hline
\hline
AlexNet & 62.36   & 58.62  & 90.00 \tabularnewline
\hline
GoogLeNet & 6.01   & 59.97  & 91.10 \tabularnewline
\hline
SqueezeNet  & 0.74   & 53.34  & 87.97 \tabularnewline
\hline
NU-LiteNet-A  & 0.27   & 66.67  & 94.07 \tabularnewline
\hline
NU-LiteNet-B  & 0.93  & 69.58 & 94.65 \tabularnewline
\hline
\end{tabular}
\end{table}

From Table 2 and Table 3, it can be observed that NU-LiteNet-A used the lowest number of parameters. NU-LiteNet-B provided the highest accuracy, while the number of parameters obtained was about three times higher than that of NU-LiteNet-A.

\begin{figure}[b]
\centering
\begin{multicols}{2}
    \includegraphics[width=\linewidth]{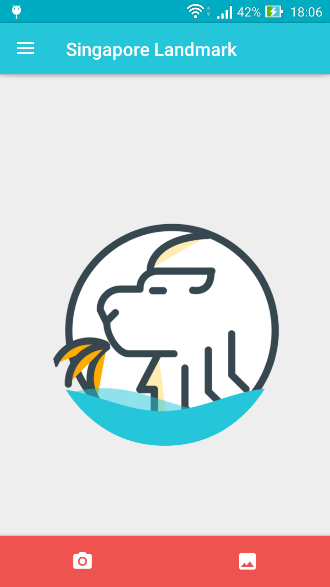}
    \includegraphics[width=\linewidth]{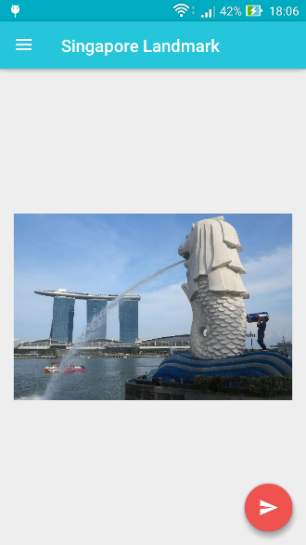}
    \end{multicols}
\caption{Snapshots from the landmark-recognition program on a smartphone with Android: (left) the first page, and (right) the query image taken by the device.}\label{fig4}
\end{figure}

\subsection{Application for Landmark Recognition on Android}

\begin{table}[t]
\centering{}\caption{EXECUTION TIME AND MODEL SIZE OBTAINED BY RECOGNI-TION ON SMARTPHONE.}
\begin{tabular}{|c|c|c|c|}
\hline
\multirow{2}{*}{Model} & Image size  & Execution time  & Model size \tabularnewline
 & (pixels) & (ms/image) & (MB)\tabularnewline
\hline
\hline
AlexNet & $1620\times1080$  & 1038  & 217 \tabularnewline
\hline
GoogLeNet & $1620\times1080$  & 1244  & 23 \tabularnewline
\hline
SqueezeNet & $1620\times1080$ & 773  & 2.86 \tabularnewline
\hline
NU-LiteNet-A &  $1620\times1080$ & 637  & 1.07 \tabularnewline
\hline
NU-LiteNet-B & $1620\times1080$ & 706 & 3.6 \tabularnewline
\hline
\end{tabular}
\end{table}

For the development of an application on smartphones using Android, the trained models were utilized for landmark recognition. The processing time and model size (the space required to store the model on a smartphone) were measured. Table 4 shows the result for processing of an input image of size $1620\times1080$ pixels. The top three models that required the lowest pro-cessing time were NU-LiteNet-A (637 ms), NU-LiteNet-B (706 ms), and SqueezeNet (773 ms). The top three mod-els that had the smallest model size were NU-LiteNet-A (1.07 MB), SqueezeNet (2.86 MB), and NU-LiteNet-B (3.6 MB). From this result, it can be observed that NU-LiteNet-A was the most effective model in terms of processing time as well as model size: 637 ms per image and 1.07 MB respectively.

Fig. 4 and 5 show snapshots of the application of mobile landmark recognition on a smartphone. The recognition function can be used in the off-line mode, in which the on-device recognition module is implemented. The user can take a picture and start the process of recognition of the landmark using the phone. The retrieved data are the name and probability score of the predicted landmark class. There are also menus for history and event that can be used to retrieve the complete information about the landmark from the web (Wikipedia) if the phone is connected to the internet. The event menu shows the information about the event currently shown at the actual are around the landmark. This information can be used to advertise the landmark to tourists.

\begin{figure}[h]
\centering
\begin{multicols}{2}
    \includegraphics[width=\linewidth]{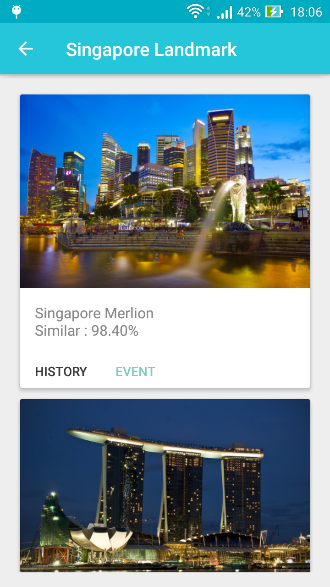}
    \includegraphics[width=\linewidth]{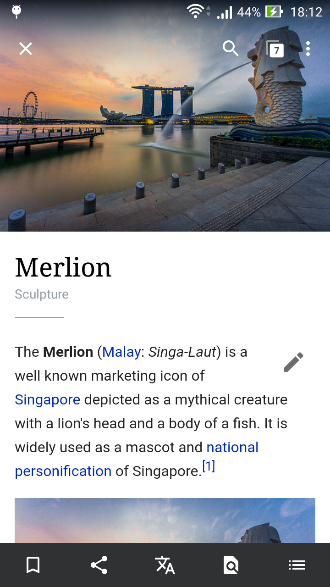}
    \end{multicols}
\caption{Snapshots from the landmark-recognition program on a smartphone with Android: (left) the recognition result showing the landmarks with the highest similarity scores in deceasing order, and (right) the information about the landmarks from Wikipedia.}\label{fig5}
\end{figure}

\section{Conclusions}

This paper presents NU-LiteNet, which adopts the development idea of SqueezeNet to improve the network structure of the convolutional neural network (CNN). It aims to reduce model size to a degree suitable for on-device processing on a smartphone. The two versions of the proposed network were tested on Singapore land-marks and a Paris dataset, and it was determined that NU-LiteNet can reduce the model size by 2.6 times compared with SqueezeNet, and improve recognition performance. The execution time of NU-LiteNet on a smartphone is also shorter than that of other CNN models. In future work, we will continue to improve accuracy and reduce model size for large-scale image databases, such as ImageNet, and country-scale landmark databases.

\appendices
\section{IMPLEMENTATION DETAILS}

The data collected in the Singapore landmarks and Paris dataset were divided into two parts: training data and testing data. The training data for the two sets was $256\times256$ pixels. Data augmentation was done using the random crop image size of $224\times224$ pixels in a horizontal flip to switch to a more increased dataset image. An improvement to enhance the accuracy of neural networks with greater precision was developed in \cite{18} by adding Batch Normalization after Convolutions all layers as well as in \cite{11,19} to allow much higher learning rates. The problem with the Internal covariate shift of \cite{20} occurred during the data training in lower hidden layers. For the Activation function, the Linear Unit Rectified \cite{21,22} (ReLU) after all the convolutions of both NU-LiteNet-A and NU-LiteNet-B.

\begin{figure}[t]
  \centering
  \includegraphics[scale=0.45]{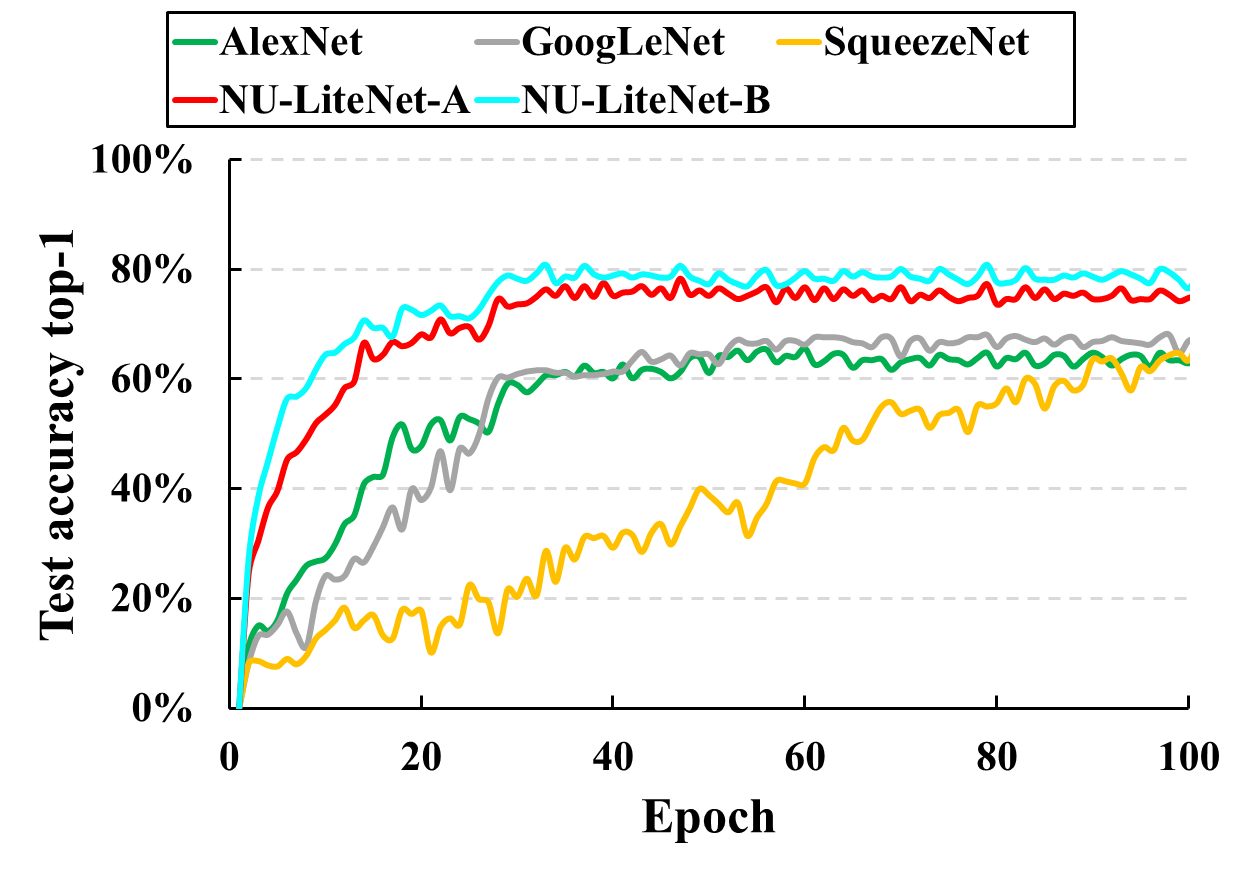}
  \caption{Top-1 accuracy vs. number of epochs; for Singapore landmarks.}\label{fig6}
\end{figure}

\begin{figure}[b]
  \centering
  \includegraphics[scale=0.45]{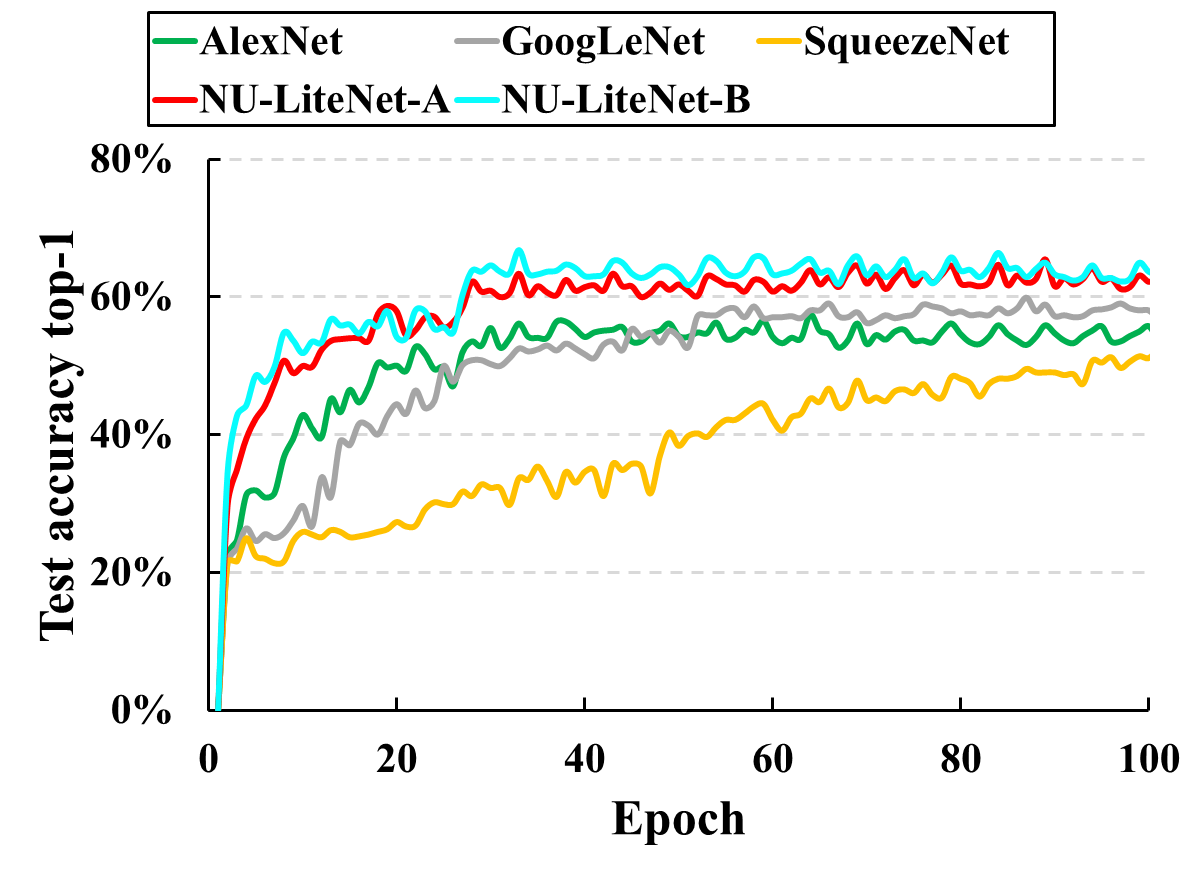}
  \caption{Top-1 accuracy vs. number of epochs; for Paris landmarks.}\label{fig7}
\end{figure}

Looking at performance top-1 accuracy of AlexNet, GoogLeNet, SqueezeNet and NU-LiteNet both versions, the training of the Singapore landmarks from epoch 1-100 was as shown in Fig. 6. Considering the accuracy of 60\%, it was observed that this model can converge before the NU-LiteNet-B at epoch 10, followed by NU-LiteNet-A at epoch 15 then GoogLeNet at epoch 29, AlexNet at epoch 34 and finally SqueezeNet at epoch 91. Considering the epoch 1-25 at learning rate (LR = 0.1) it was observed that both versions of NU-LiteNet converged better, and models AlexNet, GoogLeNet and SqueezeNet until the epoch 26 at learning rate of (LR = 0.01). The Accuracy value of both NU LiteNet is higher than all the models compared until the completion of their training. NU-LiteNet-B with 81.15\% is the highest in the series of. The model for the top1-accuracy Singapore landmarks dataset.

Similarly, when performing top-1 accuracy of AlexNet, GoogLeNet, SqueezeNet and two versions of NU-LiteNet training data set with Paris landmarks as shown during training from epoch 1-100 of Fig. 7. Considering the accuracy of 60\%, it was observed that this model can converge before the NU-LiteNet-B at epoch 28 followed by NU-LiteNet-A at epoch 29 and the models of AlexNet, GoogLeNet and SqueezeNet couldn’t converge. Accuracy is up to 60\% by the model AlexNet convergence is capped at 58.62\%, followed by model GoogLeNet which is 59.97\%, and 53.34\% on SqueezeNet. The model top1-accuracy Paris landmarks, recorded the highest accuracy for the series in NU-LiteNet-B with 69.58\%.

\ifCLASSOPTIONcaptionsoff
  \newpage
\fi



%

\bibliographystyle{IEEEtran}
\bibliography{bibliography}

%





\end{document}